\documentclass[a4paper,fleqn]{cas-sc}

\usepackage[numbers]{natbib}
\usepackage{pifont,fontawesome,svg}
\usepackage{graphicx}
\usepackage{booktabs}
\usepackage[misc]{ifsym}
\usepackage{comment}
\usepackage{subcaption}
\usepackage{multirow}
\usepackage{xcolor}
\usepackage{array}
\newcolumntype{C}[1]{>{\centering\arraybackslash}p{#1}}

\usepackage{colortbl}
\usepackage{diagbox}
\usepackage{amsmath}

\definecolor{deepblue}{RGB}{0,102,193}      % deep muted blue
\definecolor{tealgreen}{RGB}{10,190,10}     % muted teal-green
\definecolor{brownorange}{RGB}{230,20,20}     % muted brown-orange
\definecolor{lightpurple}{RGB}{201, 16, 201}
\definecolor{darkyellow}{RGB}{186,142,35}
\newcommand{\edit}[1]{\textcolor{black}{{#1}}}
\def\tsc#1{\csdef{#1}{\textsc{\lowercase{#1}}\xspace}}
\tsc{WGM}
\tsc{QE}
\begin{document}
\let\WriteBookmarks\relax
\def\floatpagepagefraction{1}
\def\textpagefraction{.001}

\shorttitle{An Empirical Benchmark of Deep Time-Series Models for Smart Meter Energy Forecasting}    

\shortauthors{Kavoosighafi et al.}  

\title [mode = title]{An Empirical Benchmark of Deep Time-Series Models for Smart Meter Energy Forecasting}  

\tnotemark[1] 

\tnotetext[1]{} 


\author[1]{Behnaz Kavoosighafi}[orcid=0000-0002-1951-7515]

\cormark[1]


\ead{behnaz.kavoosighafi@liu.se}

\ead[url]{https://liu.se/medarbetare/behka57}

\credit{Conceptualization, Data curation, Formal analysis, Investigation, Methodology, Software, Validation, Visualization, Writing – original draft, Writing – review \& editing}

\affiliation[1]{organization={Linköping University},
            addressline={Department of Science and Technology}, 
            city={Norrköping},
%          citysep={}, % Uncomment if no comma needed between city and postcode
            postcode={60174}, 
            state={Östergötland},
            country={Sweden}}

\author[2]{Maria Eidenskog}%[]
[orcid=0000-0001-9888-303X]
% Footnote of the second author

% Email id of the second author
\ead{maria.eidenskog@liu.se}

\ead[url]{https://liu.se/medarbetare/marno68}

\credit{Funding acquisition, Project administration, Writing – review \& editing}

\affiliation[2]{organization={Linköping University},
            addressline={Department of Thematic Studies}, 
            city={Linköping},
%          citysep={}, % Uncomment if no comma needed between city and postcode
            postcode={58183}, 
            state={Östergötland},
            country={Sweden}}
            
\author[2]{Wiktoria Glad}%[]
[orcid=0000-0003-3636-081X]
% Footnote of the second author

% Email id of the second author
\ead{wiktoria.glad@liu.se}

\ead[url]{https://liu.se/medarbetare/wikgl81}

\credit{Funding acquisition, Writing – review \& editing}

%\affiliation[2]{organization={Linköping University},
%            addressline={Department of Thematic Studies}, 
%            city={Linköping},
%%          citysep={}, % Uncomment if no comma needed between city and postcode
%            postcode={58183}, 
%            state={Östergötland},
%            country={Sweden}}

\author[1]{Katerina Vrotsou}%[]
[orcid=0000-0003-4761-8601]
% Footnote of the second author

% Email id of the second author
\ead{katerina.vrotsou@liu.se}

\ead[url]{https://liu.se/medarbetare/katvr62}

\credit{Conceptualization, Funding acquisition, Supervision, Resources, Writing – review \& editing}

%\affiliation[1]{organization={Linköping University},
%            addressline={Department of Science and Technology}, 
%            city={Norrköping},
%%          citysep={}, % Uncomment if no comma needed between city and postcode
%            postcode={60174}, 
%            state={Östergötland},
%            country={Sweden}}

\cortext[1]{Corresponding author}



\begin{abstract}
Accurate forecasting of energy consumption is important for the efficient operation of power systems, with direct implications for operational costs, energy management, and system maintenance. Due to the availability of extensive high-resolution consumption data from smart meters, data-driven methods have been used for short-term and long-term forecasting. However, their comparative performance on real-world smart meter data is still not well studied. In this paper, we present an empirical benchmark of nine modern deep learning models for time-series forecasting, including linear, MLP-based, convolutional, and Transformer architectures. We evaluate these models on two publicly available smart meter datasets. Our analysis focuses on three factors that strongly affect forecasting performance: the length of historical input, the prediction horizon, and the choice of model architecture. We show that extending the historical context improves accuracy, but only up to a saturation point, after which additional input provides limited benefit. In contrast, accuracy decreases as the prediction horizon increases. We also investigate the trade-off between prediction accuracy and computational complexity, and assess the statistical significance and practical magnitude of performance differences across models. Our results show that deep learning models consistently outperform classical baselines, while lightweight architectures achieve relatively similar performance at significantly lower computational cost. Additionally, architectural differences only become meaningful at longer forecasting horizons and on more heterogeneous datasets. \edit{Finally, a subgroup analysis across geodemographic and household categories shows that model choice has limited impact for most population segments, with attention-based architectures appearing to offer an advantage on the most under-represented or behaviorally irregular groups, though this advantage is not statistically robust given the small size of the affected subgroups and does not generalize consistently across datasets.} This benchmark provides practical guidance for selecting forecasting models in smart meter energy applications.
\end{abstract}


\begin{highlights}
\item We benchmark nine deep learning forecasting models on two high-resolution real-world smart meter datasets.
\item We show that forecasting accuracy improves with longer historical context only up to a saturation point, while accuracy consistently declines as the prediction horizon increases.
\item We demonstrate that lightweight architectures achieve competitive accuracy at lower computational cost, and that architectural differences become significant mainly at longer horizons and on more heterogeneous data.
\item \edit{We show, through a subgroup analysis, that Transformer-based models appear to offer an advantage on the most under-represented population segments, but this pattern is not statistically robust and does not hold consistently across datasets.}
\end{highlights}

% Keywords

\maketitle

% Main text
\section{Introduction}

Accurate forecasting of residential energy consumption is of great importance in modern energy systems, which require reliable hourly or daily load predictions to support demand response, infrastructure planning and investment, and environmental management~\cite{Wang2019}. Over the last few decades, smart meters have been deployed worldwide to record electricity, water, and other forms of consumption with higher accuracy and fine temporal resolution. As a result, many datasets containing detailed consumption information have become available, often accompanied by auxiliary factors such as weather and demographics that influence consumption patterns~\cite{Bensalah2025,Klemenjak2019,Kim2022}. These datasets provide valuable opportunities for data-driven modeling of consumption dynamics and forecasting approaches that capture complex temporal patterns and behavioral variability. At the same time, time-series forecasting has been experiencing significant progress through the adoption of machine learning techniques~\cite{Casolaro2023,Kim2025,Kong2025}. Whereas classical statistical models perform well at modeling trends and seasonality~\cite{Box1976}, neural networks are better suited for capturing non-linear temporal dependencies. Models based on convolutional networks, recurrent neural networks, and more recently transformer architectures have demonstrated strong predictive capabilities across many forecasting tasks~\cite{Scinet,Patchtst,Crossformer}. However, simpler approaches such as linear models and Multi-Layer Perceptrons (MLPs) have also been proposed to reduce computational costs while still delivering high accuracy~\cite{Dlinear,Tsmixer,Lightts}, raising the question of whether increasingly complex architectures consistently outperform simpler models when forecasting structured signals such as energy consumption.

In addition to architectural complexity, there are several other factors that directly influence the performance of forecasting models, including the length of future predictions, typically referred to as the prediction horizon, and the amount of past observations provided as input, known as the historical context. As the prediction horizon increases, forecasting becomes more challenging because models must extrapolate patterns further beyond the observed input window, which increases uncertainty and amplifies modeling errors. This is a fundamental challenge in long-horizon forecasting and has motivated the development of many specialised forecasting architectures~\cite{Lim2021,Kim2025}. 

Another important yet often overlooked aspect concerns how model performance changes across different population segments. Many studies focus mainly on aggregate predictive accuracy, without scrutinizing whether forecasting models perform similarly across diverse groups. Such analyses are important because household energy consumption is inherently heterogeneous and reflects differences in socioeconomic characteristics, occupancy patterns, and building types, among other factors~\cite{Chen2025}. %Analysing subgroup performance can therefore provide a more comprehensive understanding of model robustness and practical applicability. 
In this work, we examine forecasting performance across different socioeconomic categories to assess whether models generalise consistently across population groups.
Moreover, the amount of historical context available to a model can influence its forecasting performance~\cite{Tang2026}. While it is intuitively expected that providing more historical observations improves forecasting accuracy, this effect can change across model architectures and datasets. Some models benefit substantially from longer input sequences, whereas others may reach a performance plateau once seasonal patterns have been captured. Hence, understanding how different models exploit historical context is important when designing forecasting systems for real-world applications.

While several studies benchmark forecasting architectures on general time-series datasets, comparable evaluations on high-resolution smart meter energy data are very limited~\cite{Patchtst,Scinet,Crossformer,Tslib}. \edit{A few recent studies have begun to fill this gap, benchmarking foundation models~\cite{Meyer2025} or comparing deep learning architectures such as NBEATS, DLinear, and FEDformer~\cite{Maragkos2025} on household-level electricity data, but without jointly examining the effect of historical context length and prediction horizon, reporting an accuracy-efficiency trade-off, or assessing performance across geodemographic or building-type subgroups.} Household energy consumption data also show unique characteristics, including strong periodicity, behavioral variability, and diversity across households, which are significantly different from the aggregated datasets~\cite{Matdaut2017}. In this study, we present an empirical evaluation of forecasting algorithms for short- and long-horizon smart meter energy data and investigate how different architectures respond to increasing forecasting difficulty with respect to historical context length and prediction horizon. We evaluate nine representative models, \edit{including linear, MLP-based, convolutional, and Transformer-based designs, alongside classical statistical baselines}, on two publicly available smart meter datasets. The datasets differ in geographic coverage, temporal resolution, and consumption characteristics, which enables a broader evaluation of model behavior. \edit{Unlike prior time series benchmarks evaluated on general purpose research datasets, this study jointly examines model architecture, historical context length, prediction horizon, computational cost, statistical and practical significance, and geodemographic subgroup fairness on real, high resolution smart meter data, providing evidence-based guidance for model selection in operational energy forecasting settings.} Our main contributions are summarized as follows:

\begin{itemize}
    \item An empirical benchmark of nine forecasting architectures, including linear, MLP-based, convolutional, and Transformer-based models, alongside classical baselines (SARIMA, na\"ive seasonal) on two real-world smart meter datasets.
    \item An exploratory analysis of the datasets, including consumption patterns, temporal behavior, and the distribution of demographic subgroups.
    \item An analysis of the trade-off between predictive accuracy and computational efficiency, together with an investigation of how historical context length and prediction horizon influence forecasting performance.
    \item A subgroup analysis across geodemographic categories, highlighting potential disparities in predictive performance across population segments.
\end{itemize}

\noindent The code used in this study is publicly available at\edit{~\url{https://github.com/behnazkavoosi/Energy-Time-Series-Library}}.%www.dummy.url}.
%- how do we differ from the previous benchmarks/surveys?
%- how geodemographic info is related to forecasting?
%- why we chose these datasets and why these models?
%- why random selection? why not some other strategies?

\section{Related Work}

\subsection{Time-Series Forecasting}

Time-series forecasting concerns the estimation of future observations based on historical data and is an active research area in machine learning, statistics, and related disciplines~\cite{Box1990}. Given the inherent complexity of temporal dynamics, forecasting models aim to learn the underlying structure of the series, such as trend and seasonality, in order to extract informative representations that result in accurate predictions. %Depending on the number of input variables, forecasting methods are typically categorised as \emph{univariate} or \emph{multivariate}~\cite{Chatfield2000}. Univariate forecasting considers a single target variable, for example, predicting future energy consumption only from past consumption values~\cite{Newbold1974}. In contrast, multivariate forecasting incorporates additional covariates, such as temperature or humidity, together with historical consumption data. Although multivariate models can exploit inter-variable dependencies and often yield improved predictive performance, this advantage comes at the cost of increased model complexity and higher data requirements.

Forecasting approaches can be grouped according to the prediction horizon into short-term and long-term forecasting. The definition of a long prediction horizon depends on the temporal resolution of the data. For example, one week corresponds to 168 steps in hourly data but only seven in daily data. %There is no universally accepted definition of a long horizon; rather, it is determined relative to the temporal resolution of the data and the dynamics of the underlying system. For instance, forecasting one week ahead corresponds to 168 time steps in hourly data and may be regarded as a long-horizon prediction, whereas the same calendar duration represents only seven time steps in daily data and may therefore be considered a short-horizon prediction. 
Long-term forecasting is usually applied in infrastructure planning, market analysis, maintenance, and scheduling. However, predictive uncertainty generally increases with forecast length because of error accumulation and the gradual loss of reliable contextual information~\cite{Li2023}. Short-term forecasting, by contrast, typically achieves higher accuracy and is often employed for operational decision-making and demand response~\cite{Hagan1987}.

Forecasting can be performed using classical statistical methods, machine learning approaches, and deep learning models. Statistical models estimate future values using mathematical formulations that approximate the behavior of historical data~\cite{Gooijer2006}. Well-known examples include Auto-Regressive (AR), Moving Average (MA), and Auto-Regressive Integrated Moving Average (ARIMA)~\cite{Box1976}, which are often effective for short-term forecasting but struggle with complex or highly non-linear data. Machine learning models aim to improve predictive performance through supervised learning~\cite{Ahmed2010}, with methods such as Random Forest~\cite{Breiman2001}, eXtreme Gradient Boosting (XGBoost)~\cite{Chen2016}, and Support Vector Machines (SVM)~\cite{Cortes1995}. More recently, deep learning approaches, including Temporal Convolutional Networks (TCNs)~\cite{Bai2018}, Recurrent Neural Networks (RNNs)~\cite{Elman1990}, and Transformers~\cite{Vaswani2017}, have advanced forecasting by learning complex temporal dependencies and interactions in multivariate data. This flexibility, however, oftentimes comes at the cost of higher computational and data requirements. For comprehensive overviews of forecasting methods, we refer the reader to recent surveys~\cite{Casolaro2023,Kim2025,Kong2025,Lim2021}. %,Masini2023,Torres2020}.  

\subsection{Energy Consumption Forecasting}

Energy and load consumption forecasting has been an active area of research for many years and supports applications ranging from long-term strategic planning and renewable energy integration~\cite{Wang2019} to building energy system management. In this context, energy data refers to various forms of consumption, including electricity, heating, and water. Predictions may be done at an aggregated scale, estimating total demand for a region or city~\cite{Khan2021}, or at a granular scale, focusing on individual households, buildings, or even specific appliances~\cite{Matdaut2017}. Aggregated forecasts are mainly used for urban energy management and cost control, whereas granular predictions are more relevant for engineering, maintenance, and operational optimization~\cite{Somu2021}. %In energy applications, prediction horizons vary depending on the use case. Short-horizon forecasts are typically associated with operational tasks, such as HVAC optimisation, while long-horizon forecasts inform strategic decisions, including renovation and retrofit planning.

Approaches developed for general time-series analysis are commonly applied in this domain. In the univariate setting, models use historical consumption values as input to generate predictions~\cite{Saab2001}. In the multivariate scenario, however, additional covariates such as weather conditions, occupancy patterns, and time-of-day or day-of-week indicators are incorporated to enhance performance~\cite{Gonzalez2019}. For daily-resolution data, derived features such as the daily mean, minimum, maximum, and standard deviation can also be included. For a comprehensive overview of methods in energy consumption forecasting, we refer the reader to~\cite{Ahmad2014,Deb2017,Wei2019}. Details of the deep learning-based models used in our experiments are provided in Section~\ref{sec:models}.

\section{Benchmarking}
Energy consumption forecasting plays an important role in reducing operational costs and improving energy efficiency. In this study, we benchmark a range of forecasting algorithms on two different energy datasets in order to examine the following research questions:
\begin{enumerate}
    \item To what extent does model choice influence predictive performance on high-resolution smart meter data?
    \item How does the length of historical context affect forecasting accuracy across model families?
    \item How does predictive performance degrade as the forecast horizon increases, and do different model classes exhibit distinct robustness to long-horizon prediction?
    \item Do forecasting models perform similarly across geodemographic population segments, or do certain architectures show advantages for specific groups?

\end{enumerate}
Our main goal is to evaluate modeling decisions under realistic data conditions. We therefore begin with an analysis of datasets to characterize their statistical properties, and then describe the forecasting models and experimental protocol used in the benchmark.

\begin{figure}[t]
    \centering
    \begin{subfigure}[b]{0.8\linewidth}
        \centering
        \includegraphics[width=\linewidth]{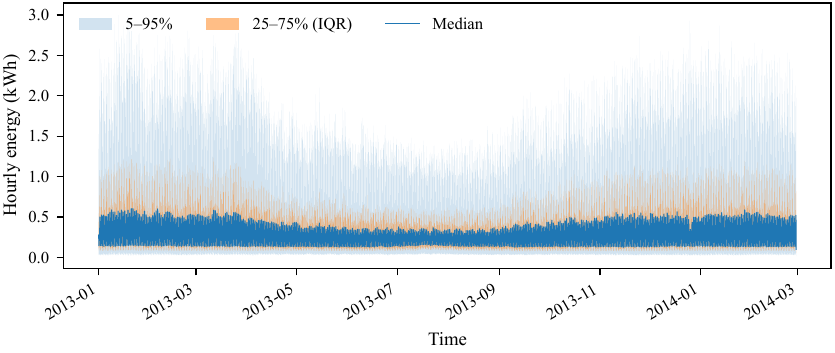}
        \caption{}
        \label{fig:lcl-a}
    \end{subfigure}
    \vspace{0.4em}

    \begin{subfigure}[b]{0.8\linewidth}
        \centering
        \includegraphics[width=\linewidth]{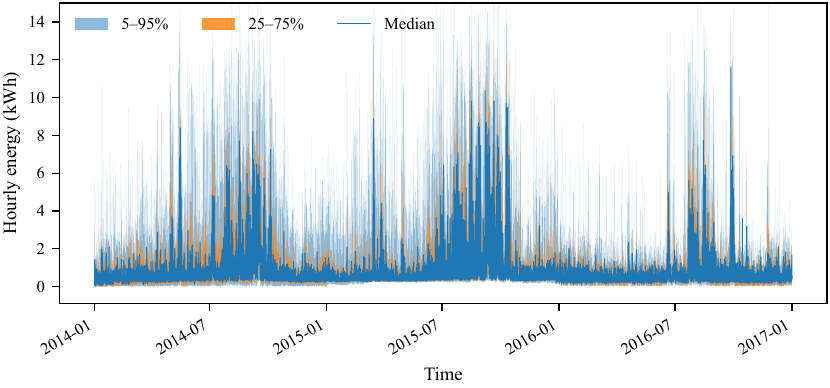}
        \caption{}
        \label{fig:pecan-a}
    \end{subfigure}
    \vspace{0.4em}

    \begin{subfigure}[b]{0.47\linewidth}
        \centering
        \includegraphics[width=\linewidth]{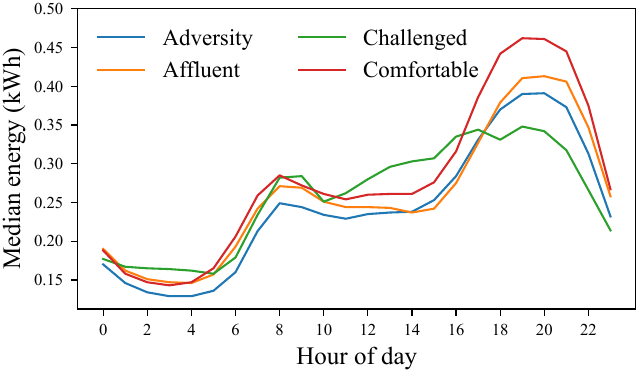}
        \caption{}
        \label{fig:lcl-b}
    \end{subfigure}
    \hfill
    \begin{subfigure}[b]{0.44\linewidth}
        \centering
        \includegraphics[width=\linewidth]{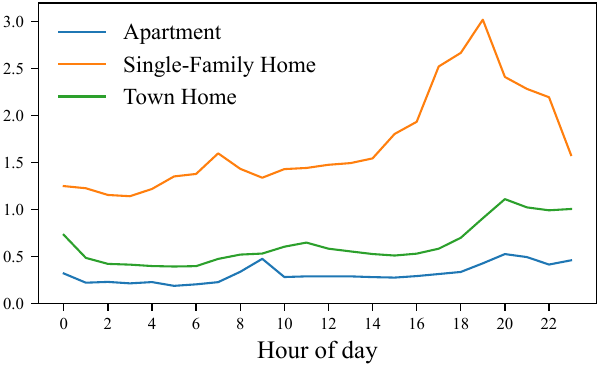}
        \caption{}
        \label{fig:pecan-b}
    \end{subfigure}
    \vspace{0.4em}

    \caption{Hourly electricity consumption patterns in the LCL and Pecan Street datasets. (a) Population-level hourly envelope for the LCL dataset, showing the median and 25--75\% and 5--95\% quantile bands across households. (b) Population-level hourly envelope for the Pecan Street dataset, showing the median and 25--75\% and 5--95\% quantile bands across households over the period 2014--2017. (c) Median intraday profiles by ACORN socioeconomic group for the LCL dataset. (d) Median intraday profiles by building type for the Pecan Street dataset.}
    \label{fig:lcl}
\end{figure}

\subsection{Datasets}\label{sec:datasets}
We utilized two publicly available smart meter electricity datasets for our experiments: (1) the \emph{Low-Carbon London (LCL)}\footnote{https://data.london.gov.uk/dataset/smartmeter-energy-consumption-data-in-london-households-vqm0d/} dataset and (2) the \emph{Pecan Street}\footnote{https://www.pecanstreet.org/dataport/} dataset. 
The LCL dataset contains electricity consumption records from 5\,561 households in London, collected at a half-hourly resolution over 829 days, spanning November 2011 to February 2014. In addition to consumption data, the dataset is accompanied by survey information that categorizes households according to geodemographic characteristics, including income, education level, and age, among other attributes. However, the original survey responses are not directly accessible. Instead, households are assigned ACORN categories\footnote{https://acorn.caci.co.uk/how-acorn-works/}, which provide aggregated geodemographic labels and therefore do not permit direct mapping to the underlying survey variables. 
In the Pecan Street dataset, measurements were recorded at 15-minute resolution, mainly over 2013--2017, with additional data extending into 2018--2019 for a subset of households. The dataset includes 73 households from New York, California, and Austin in the United States and provides appliance- or circuit-level electricity measurements. In this work, we computed gross household load by summing \edit{all recorded circuit-level consumption channels for each household, excluding grid import/export, solar generation, battery storage, and voltage-only channels.} Alongside the data, additional variables such as year of construction, housing type, and house size are also collected to complement the energy data.

Figure~\ref{fig:lcl} illustrates the hourly electricity consumption patterns in the LCL and Pecan Street datasets. For the LCL dataset, the original half-hourly measurements were resampled to hourly resolution for these visualizations and throughout the benchmark in order to reduce computational complexity. Additionally, the analysis is limited to the period January 2013 to March 2014 due to incomplete records for many households in earlier years and ensure a more consistent evaluation period. For the Pecan Street dataset, the original 15-minute resolution was retained for all experiments to assess model performance at a finer temporal granularity and to compensate for the limited number of households through a higher number of time steps per series; for visualization purposes, however, the 15-minute data were aggregated to hourly resolution. Figures~\ref{fig:lcl-a} and~\ref{fig:pecan-a} present the population-level hourly envelope over the experimental period for the LCL and Pecan Street datasets, respectively. The median, together with the 25--75\% and 5--95\% quantile bands, is computed across households at each timestamp. The resulting envelopes indicate obvious annual seasonality, with higher consumption during colder months and lower levels in summer for the LCL dataset, and the opposite for the Pecan Street dataset, which reflects climate differences between the two datasets. Note that the Pecan Street dataset exhibits higher consumption magnitudes, with peak hourly values going beyond 10~kWh compared to approximately 3~kWh in the LCL dataset, as well as greater household-level variability, as confirmed by the wider quantile bands and more extreme spikes in the population envelope. Figure~\ref{fig:lcl-b} compares median intraday profiles across four geodemographic ACORN categories within the LCL dataset. Households in the \emph{Affluent} and \emph{Comfortable} groups show higher consumption than those in the \emph{Adversity} and \emph{Challenged} groups. Differences are most visible during peak hours, while night-time consumption is relatively similar across categories. We also note that the dataset is socio-economically imbalanced, including 39.4\% \emph{Affluent}, 31.9\% \emph{Adversity}, 27.4\% \emph{Comfortable}, and only 1.3\% \emph{Challenged} households. Results for the \emph{Challenged} group should therefore be interpreted with caution due to limited representation. Additionally, Figure~\ref{fig:pecan-b} illustrates median intraday patterns across building types in the Pecan Street dataset. In terms of peak consumption, single-family homes show considerably higher consumption than town homes and apartments, although the overall profile shape is similar across building types. As with the LCL dataset, the Pecan Street dataset is imbalanced with respect to building type: approximately 70\% of households are single-family homes, 20\% town homes, and 10\% apartments.

\begin{table}
\centering
\caption{Comparison of forecasting models used in our benchmark.}
\label{tab:model_comparison}
\begin{tabular}{lccc}
\toprule
Model & Architecture & Core Mechanism & Complexity \\
\midrule
DLinear~\cite{Dlinear}     & Linear model & Series decomposition & Very low \\
LightTS~\cite{Lightts}     & MLP-based & 	Interval and continuous sampling & Low \\
TSMixer~\cite{Tsmixer}     & MLP-based & Time-channel mixing & Low \\
FiLM~\cite{Film}        & Memory-based & Frequency-domain denoising & Medium \\
SCINet~\cite{Scinet}      & CNN-based & Recursive downsampling  & Medium \\
TiDE~\cite{Tide}        & MLP-based & Residual encoder-decoder  & Medium \\
Crossformer~\cite{Crossformer} & Transformer & Two-stage attention & High \\
PatchTST~\cite{Patchtst}    & Transformer & Patching and channel independence  & High \\
iTransformer~\cite{iTransformer} & Transformer   & Inverted attention across variates & High   \\
\bottomrule
\end{tabular}
\end{table}

% why these specific models were selected
\subsection{Models} \label{sec:models}

A wide range of algorithms has been proposed for both long- and short-horizon time-series forecasting, including applications to smart meter energy data. In this study, we selected nine models, ranging from simpler linear approaches to more complex and computationally intensive Transformer-based architectures. The models were chosen to cover the major architectural families in recent time-series forecasting literature, with each model representing a distinct design principle, including linear decomposition, interval and continuous sampling, time-channel mixing, residual encoder--decoder, multi-scale convolution, frequency-domain encoding, and various attention-based mechanisms, so that the benchmark captures a representative and diverse set of modern approaches without redundancy. \edit{From a deployment perspective, DLinear, LightTS, and TSMixer's low computational cost is relevant for utility scale forecasting across thousands of meters, while the cross variate modeling capacity of PatchTST, iTransformer, and Crossformer is more relevant to multivariate settings incorporating covariates such as weather or occupancy.} Table~\ref{tab:model_comparison} summaries the models included in our benchmark, together with their architectural design, core mechanisms, and relative theoretical computational complexity. 

At the simpler end, \textbf{DLinear}~\cite{Dlinear} is a decomposition-based linear model that separates the input into trend and seasonal components using a moving average filter and applies dedicated linear layers to each part. \textbf{LightTS}~\cite{Lightts} and \textbf{TSMixer}~\cite{Tsmixer} adopt MLP-based designs: LightTS combines continuous and interval sampling to capture local and global patterns efficiently, while TSMixer alternates between time-mixing and feature-mixing MLP layers to model temporal dependencies and cross-variate interactions. Similarly, \textbf{TiDE}~\cite{Tide} employs a dense MLP-based encoder–decoder architecture with residual connections to incorporate historical observations and dynamic covariates. Moving towards convolutional and hybrid designs, \textbf{SCINet}~\cite{Scinet} utilises a recursive binary tree structure within an encoder–decoder framework, where sub-sequences interact through convolutional layers to capture multi-scale temporal dependencies. \textbf{FiLM}~\cite{Film} integrates Legendre Projection Units to encode long historical context together with Fourier-based enhancement modules to model frequency-domain characteristics. Finally, the Transformer-based models \textbf{Crossformer}~\cite{Crossformer}, \textbf{PatchTST}~\cite{Patchtst}, and \textbf{iTransformer}~\cite{iTransformer} leverage attention mechanisms: Crossformer applies a two-stage attention scheme to capture cross-time and cross-dimension dependencies, whereas PatchTST divides the input sequence into fixed-length temporal segments and processes them using channel-independent Transformer encoders, modeling each variable separately while sharing parameters across channels. iTransformer embeds the entire time series of each variable into a single token and applies self-attention across variables, making it suitable for capturing multivariate dependencies. For a comprehensive analysis of each architecture, we refer the reader to the original publications.

\subsection{Experimental Setup}

Our pipeline consisted of three stages: data pre-processing, model training, and evaluation (see Section~\ref{sec:disucssion}). For LCL, we randomly selected 560 out of 5\,561 households to reduce computational cost while ensuring representation across all ACORN groups. We limited the analysis to January 2013--March 2014 due to data completeness and resampled the original half-hourly measurements to hourly resolution. For the Pecan Street dataset, we used all 73 households at the original 15-minute resolution. We pre-processed both datasets using standard strategies, including timestamp parsing, removal of invalid entries, and interpolation of missing values where necessary. \edit{Specifically, values that fail numeric parsing are treated as missing and filled by linear interpolation using neighboring valid readings in both directions. Any values still missing after interpolation, for example at the start or end of a household's series where interpolation has no bound on one side, are filled with that household's mean consumption.} For LCL, we applied global standardization using the mean and standard deviation computed from the training portion across all households in order to maintain inter-household magnitude differences. For Pecan Street, we performed per-household normalization, fitting scaling parameters independently on each household’s training segment. \edit{This choice was necessary because Pecan Street households exhibit significantly larger scale heterogeneity than LCL, while comprising far fewer series. As a result, a single global scaling statistic could be a poor fit for many individual households and dominate the loss by the highest-consumption households. Note that, as a consequence of this difference in pre-processing, absolute error magnitudes are not directly comparable between the two datasets.} Once normalization was done, sliding windows were then constructed and split chronologically into 80\% training, 10\% validation, and 10\% test sets. Note that all normalization statistics were computed on the training data to prevent data leakage. %In addition to the univariate setting, we evaluated multivariate forecasting on the LCL dataset, with energy consumption as the target variable and temperature, dew point, and humidity\footnote{https://www.kaggle.com/datasets/jeanmidev/smart-meters-in-london} as exogenous covariates, following similar pre-processing and scaling procedures.

For model training, we used the Time Series Library (TSLib)~\cite{Tslib}, which provides implementations of 30 forecasting models. We followed the original implementations and trained all models under a consistent sliding-window forecasting setup to ensure fair comparison. From each household time series, we extracted input segments of fixed length (\verb!seq_len!) and trained the models to predict the next \verb!pred_len! time steps. For encoder–decoder architectures, an additional overlap segment (\verb!label_len!) was used to condition the decoder during forecasting. The same window lengths, chronological data splits, and pre-processing procedures were applied across all models so that performance differences reflect architectural designs.
We trained the models, listed in Table~\ref{tab:model_comparison}, for a fixed budget of 5 epochs with a learning rate of 10$^{-4}$ and a batch size of 64 for LCL and 128 for Pecan Street, using the Adam optimizer~\cite{Kingma2014} and Mean Squared Error (MSE) as the loss function. \edit{While the training budget, learning rate, batch size, optimizer, and loss function were held fixed across all nine models, each model used its default architecture-specific hyperparameters (e.g., number of layers, hidden dimensions, attention heads) as implemented in TSLib. We therefore report our results as a fixed-compute comparison under each model's default architecture configuration, rather than a comparison under independently optimized configurations.} Note that in our preliminary experiments, we extended training to 10 epochs but observed no improvement in validation performance, which suggests that models had largely stabilized within the selected training budget. All experiments were conducted on an NVIDIA RTX 5090 GPU. 

\begin{table}[t]
\centering
\caption{Forecast accuracy of the deep learning models and classical baselines on the LCL dataset at $\texttt{seq\_len} = 24$ and $\texttt{pred\_len} = 24$. The best value in each column is \textbf{bold}.}
\label{tab:baseline_comparison}
\setlength{\tabcolsep}{5pt}
\begin{tabular}{lccccccc}
\toprule
Model
  & MSE $\downarrow$
  & MAE $\downarrow$
  & MAPE $\downarrow$
  & $R^2$ $\uparrow$
  & $r$ $\uparrow$
  & $\cos$ $\uparrow$
  & FD $\downarrow$ \\
\midrule

\multicolumn{7}{l}{\textit{Classical baselines}} \\
Na\"ive seasonal   & 0.240 & 0.241 & \textbf{0.721} & 0.391 & 0.697 & 0.792 & 0.712 \\
SARIMA           & 0.196 & 0.246 & 0.991 & 0.504 & 0.739 & 0.820 & \textbf{0.631} \\
\midrule

\multicolumn{7}{l}{\textit{Deep learning models}} \\
DLinear          & 0.177 & 0.227 & 1.310 & 0.553 & 0.744 & 0.857 & 0.960 \\
LightTS          & 0.166 & 0.216 & 1.226 & 0.579 & 0.761 & 0.863 & 0.982 \\
TSMixer          & 0.168 & 0.217 & 1.203 & 0.576 & 0.760 & 0.861 & 0.988 \\
FiLM             & 0.177 & 0.221 & 0.863 & 0.551 & 0.743 & 0.857 & 0.969 \\
SCINet           & 0.174 & 0.218 & 0.854 & 0.560 & 0.749 & 0.858 & 0.974 \\
TiDE             & 0.177 & 0.221 & 0.863 & 0.551 & 0.743 & 0.857 & 0.968 \\
Crossformer      & \textbf{0.165} & 0.217 & 1.082 & \textbf{0.582} & \textbf{0.763} & 0.862 & 0.973 \\
PatchTST         & 0.167 & \textbf{0.214} & 0.915 & 0.577 & 0.760 & 0.862 & 0.991 \\
iTransformer     & 0.166 & 0.218 & 1.343 & 0.581 & 0.762 & \textbf{0.863} & 0.953 \\
\bottomrule
\end{tabular}
\end{table}

\section{Results}
\label{sec:disucssion}

We evaluate all models across multiple prediction horizons and historical context lengths using global test metrics, including Mean Squared Error (MSE), Mean Absolute Error (MAE), Mean Absolute Percentage Error (MAPE), and the coefficient of determination ($R^2$). \edit{Note that all reported metrics are computed after inverse-transforming model predictions and ground-truth values back to their original physical scale (kWh).} While MSE, MAE, and MAPE quantify point-wise prediction error, $R^2$ captures the proportion of variance in the target explained by the forecasts. \edit{MAPE is computed with a small epsilon added to the denominator to avoid division by zero.} To assess structural similarity between predicted and ground-truth trajectories, we additionally report Fréchet distance (FD), cosine similarity, and Pearson's correlation coefficient ($r$). \edit{Fréchet distance and cosine similarity reflect how well a model captures the shape and timing of consumption, which is relevant to demand-response scheduling. Pearson correlation, on the other hand, captures whether relative fluctuations are tracked even when magnitude is off, useful for ranking households by relative load.} Computational efficiency is evaluated in terms of training \edit{and inference} time. In addition to aggregate metrics, we conduct subgroup analyses for relevant categories, such as ACORN groups in the LCL dataset. \edit{We also analyze how model-specific mechanisms account for performance differences between models under different conditions.} Note that unless otherwise stated, all results reported in this section correspond to the LCL dataset; full results for both datasets are provided in the Supplementary Material.

To establish a baseline, we compare the deep learning models against two classical baselines: Na\"ive seasonal and Seasonal ARIMA (SARIMA)~\cite{Box1976}. The baselines were evaluated on the same test windows as the deep learning models to ensure a fair comparison. The na\"ive seasonal baseline predicts the next \texttt{pred\_len} steps by repeating the observations from one seasonal cycle prior to the forecast origin. SARIMA models were fitted independently per household using AIC-based order selection from a set of candidate orders, with a seasonal period of 24 hours for the LCL dataset and 96 steps (one day) for the Pecan Street dataset. A context window of three seasonal cycles was used for Kalman filter initialization, and all metrics were computed in the same normalized space as the deep learning outputs to ensure comparability. Table~\ref{tab:baseline_comparison} compares all models at \texttt{seq\_len} = 24 and \texttt{pred\_len} = 24 for the LCL dataset. Deep learning models clearly outperform both classical baselines across the primary accuracy metrics (MSE, MAE, $R^2$, Pearson's $r$, and cosine similarity), confirming that learned temporal representations capture consumption dynamics more effectively than statistical approaches. This advantage is even more considerable on the Pecan Street dataset, where both baselines result in negative $R^2$ values, which shows their inability to handle the higher household-level variability of this dataset. Interestingly, the na\"ive seasonal and SARIMA achieve the lowest MAPE and Fréchet distance, respectively, compared to most deep learning models; however, these are metric-specific artifacts: the na\"ive seasonal baseline benefits from MAPE's sensitivity to small absolute values in low-consumption periods, while SARIMA's lower Fréchet distance reflects the inherent smoothness of autoregressive predictions.

\begin{table}
\centering
\caption{Best-performing forecasting model (lowest composite score) for each combination of input sequence length (\texttt{seq\_len}) and prediction horizon (\texttt{pred\_len}) for the LCL dataset. }
\label{tab:pred_vs_seq_lcl}
\begin{tabular}{|c|C{1.7cm}C{1.7cm}C{1.7cm}C{1.7cm}C{1.7cm}|}
\hline
\diagbox{\texttt{pred\_len}}{\texttt{seq\_len}} & 24 & 96 & 168 & 336 & 672 \\
\hline
\hline
24 & PatchTST%, PatchTST, TSMixer, Crossformer, SCINet, DLinear, TiDE, FiLM
& iTransformer%, LightTS, TSMixer, Crossformer, DLinear, SCINet, TiDE, FiLM
& iTransformer%, TSMixer, LightTS, DLinear, TiDE, SCINet, Crossformer, FiLM
& iTransformer%, PatchTST, LightTS, DLinear, Crossformer, SCINet, TiDE, FiLM
& iTransformer%, TSMixer, LightTS, DLinear, TiDE, Crossformer, SCINet, FiLM
\\
\hline
\hline
96 & PatchTST%, PatchTST, TSMixer, Crossformer, DLinear, SCINet, TiDE, FiLM
& PatchTST%, LightTS, TSMixer, Crossformer, SCINet, DLinear, TiDE, FiLM
& iTransformer%, TSMixer, LightTS, Crossformer, DLinear, TiDE, SCINet, FiLM
& iTransformer%, LightTS, PatchTST, Crossformer, DLinear, TiDE, SCINet, FiLM
& PatchTST%, PatchTST, Crossformer, DLinear, TiDE, LightTS, SCINet, FiLM
\\
\hline
\hline
168 & TSMixer%, LightTS, TSMixer, Crossformer, DLinear, SCINet, TiDE, FiLM
& TSMixer%, PatchTST, TSMixer, Crossformer, DLinear, SCINet, TiDE, FiLM
& TSMixer%, TSMixer, PatchTST, Crossformer, DLinear, TiDE, SCINet, FiLM
& Crossformer%, Crossformer, LightTS, PatchTST, DLinear, TiDE, SCINet, FiLM
& TiDE%, TSMixer, PatchTST, LightTS, TiDE, DLinear, SCINet, FiLM
\\
\hline
\end{tabular}
\end{table}

\begin{table}
\centering
\caption{Best-performing forecasting model (lowest composite score) for each combination of input sequence length (\texttt{seq\_len}) and prediction horizon (\texttt{pred\_len}) for the Pecan Street dataset.}
\label{tab:pred_vs_seq_pecan}
\begin{tabular}{|c|C{1.8cm}C{1.8cm}C{1.8cm}C{1.8cm}|}
\hline
\diagbox{\texttt{pred\_len}}{\texttt{seq\_len}} & 96 & 384 & 672 & 1344 \\
\hline
\hline
96 & PatchTST
& TiDE
& Crossformer
& iTransformer
\\
\hline
\hline
384 & TiDE
& PatchTST
& TiDE
& iTransformer
\\
\hline
\hline
672 & PatchTST
& TiDE
& TiDE
& Crossformer
\\
\hline
\end{tabular}
\end{table}

\subsection{Accuracy-Efficiency Model Selection}

Considering that different models have different levels of architectural complexity (see Table~\ref{tab:model_comparison}), we explore the trade-off between prediction accuracy and computational cost. To this end, we introduce a composite metric $\mathcal{L} = \alpha\,\text{MAE}_n + \beta\,\text{time}_n$, where $\text{MAE}_n$ and $\text{time}_n$ are the normalized MAE and training time for a given configuration, obtained via MinMax scaling. The weights $\alpha$ and $\beta$ are set to 0.7 and 0.3, respectively, to prioritize predictive accuracy while accounting for computational cost; see Supplementary Material for a sensitivity analysis under alternative weightings. Tables~\ref{tab:pred_vs_seq_lcl} and~\ref{tab:pred_vs_seq_pecan} report the best-performing model under this metric for each combination of \texttt{seq\_len} and \texttt{pred\_len} on the LCL and Pecan Street datasets, respectively.

For the LCL dataset, at short horizons (\texttt{pred\_len} = 24), PatchTST achieves the lowest composite score at the shortest input length (\texttt{seq\_len} = 24), while iTransformer dominates across all longer input windows (\texttt{seq\_len} $\geq$ 96), suggesting that patch-based attention is most effective under limited context whereas the inverted attention mechanism of iTransformer benefits progressively from larger historical windows. At intermediate horizons (\texttt{pred\_len} = 96), PatchTST performs best at \texttt{seq\_len} $\in \{24, 96, 672\}$, while iTransformer leads at \texttt{seq\_len} $\in \{168, 336\}$. At the longest horizon (\texttt{pred\_len} = 168), the results shift towards lighter architectures: TSMixer achieves the lowest composite score for \texttt{seq\_len} $\in \{24, 96, 168\}$, Crossformer at \texttt{seq\_len} = 336, and TiDE at \texttt{seq\_len} = 672, indicating that the relative advantage of Transformer-based models diminishes at longer horizons where MLP-based and encoder-decoder architectures provide a better accuracy-efficiency trade-off. In contrast, results on the Pecan Street dataset show that PatchTST and TiDE dominate most configurations, with PatchTST leading at shorter context lengths (\texttt{seq\_len} $\in \{96, 384\}$) for \texttt{pred\_len} $\in \{96, 672\}$ and TiDE performing best across the majority of medium and long input windows, which implies that the encoder-decoder design of TiDE is better suited to handle the higher variability and finer temporal resolution of this dataset. iTransformer outperforms other models at the largest context window (\texttt{seq\_len} = 1344) for shorter prediction horizons, and Crossformer at \texttt{seq\_len} $\in \{672, 1344\}$ for the longest horizon.

These findings demonstrate that no single model performs best across all configurations, with the optimal choice depending on both the forecasting horizon and the historical context length. \edit{The sensitivity analysis in the Supplementary Material confirms that these patterns are mostly stable across alternative weightings: PatchTST and iTransformer dominate at shorter prediction horizons regardless of weighting, while lightweight models become increasingly competitive at longer horizons and more balanced weightings} DLinear, LightTS, FiLM, and SCINet do not achieve the lowest composite score under any configuration with the primary weighting scheme on either dataset. \edit{Note that model rankings on efficiency reflect architectural differences rather than hardware-specific effects, and are therefore expected to generalize across GPU platforms; however, this has not been empirically verified on edge or resource-constrained devices, where absolute training and inference times may differ substantially.}

\begin{table}
\centering
\caption{Forecasting performance on the LCL dataset across different historical context lengths ($s$, in hours) for \texttt{pred\_len} = 24. MAPE values are reported in units of $\times 10^{3}$. Here, $r$ denotes the Pearson correlation coefficient, $\cos$ the cosine similarity, $FD$ the Fréchet distance, and $tt$ and $it$ the training and inference time in minutes and milliseconds, respectively. The best result per context length is highlighted in \textbf{bold}, and the overall best result is emphasized in \underline{\textbf{bold and underlined}}.}
\label{tab:lcl-24}

\begin{tabular}{C{0.7cm} C{1.8cm} C{1.1cm} C{1.1cm} C{1.3cm} C{1.1cm} C{1.1cm} C{1.1cm} C{1.1cm} C{1.0cm} C{0.7cm}}
\toprule
$s$ & Model & MSE $\downarrow$ & MAE $\downarrow$ & MAPE $\downarrow$ & $R^2$ $\uparrow$ & $r$ $\uparrow$ & $\cos$ $\uparrow$ & FD $\downarrow$ & $tt$ $\downarrow$ & $it$ $\downarrow$ \\
\midrule

\multirow{9}{*}{24}
 & DLinear      & 0.177 & 0.227 & 1.310 & 0.553 & 0.744 & 0.857 & 0.960 & \textbf{21.82} & \textbf{0.24}\\
 & LightTS      & 0.166 & 0.216 & 1.226 & 0.579 & 0.761 & 0.863 & 0.982 & 36.51 & 0.80\\
 & TSMixer      & 0.168 & 0.217 & 1.203 & 0.576 & 0.760 & 0.861 & 0.988 & 29.75 & 0.98\\
 & FiLM         & 0.177 & 0.221 & 0.863 & 0.551 & 0.743 & 0.857 & 0.969 & 198.96 & 7.46\\
 & SCINet       & 0.174 & 0.218 & \textbf{0.854} & 0.560 & 0.749 & 0.858 & 0.974 & 307.76 & 12.11\\
 & TiDE         & 0.177 & 0.221 & 0.863 & 0.551 & 0.743 & 0.857 & 0.968 & 43.35 & 1.32\\
 & Crossformer  & \textbf{0.165} & 0.217 & 1.082 & \textbf{0.582} & \textbf{0.763} & 0.862 & 0.973 & 332.18 & 8.96\\
 & PatchTST     & 0.167 & \textbf{0.214} & 0.915 & 0.577 & 0.760 & 0.862 & 0.991 & 50.84 & 1.70 \\
 & iTransformer & 0.166 & 0.218 & 1.343 & 0.581 & 0.762 & \textbf{0.863} & \textbf{0.953} & 49.01 & 1.20\\
\midrule

\multirow{9}{*}{96}
 & DLinear      & 0.151 & 0.208 & 1.096 & 0.618 & 0.786 & 0.884 & 0.738 & \textbf{21.89} & \textbf{0.39} \\
 & LightTS      & 0.144 & 0.203 & 1.020 & 0.635 & 0.797 & 0.889 & 0.752 & 39.13 & 1.30\\
 & TSMixer      & 0.144 & 0.205 & 1.140 & 0.635 & 0.797 & \textbf{0.894} & 0.743 & 32.91 & 0.97\\
 & FiLM         & 0.151 & 0.205 & \textbf{0.863} & 0.617 & 0.786 & 0.884 & 0.738 & 334.63 & 9.88\\
 & SCINet       & 0.149 & 0.205 & 0.866 & 0.622 & 0.789 & 0.888 & \textbf{0.733} & 438.68 & 12.99\\
 & TiDE         & 0.151 & 0.206 & 0.865 & 0.617 & 0.786 & 0.884 & 0.736 & 44.37 & 1.34\\
 & Crossformer  & 0.145 & 0.201 & 0.869 & 0.632 & 0.795 & 0.886 & 0.755 & 377.66 & 8.97 \\
 & PatchTST     & \textbf{0.143} & \textbf{0.200} & 0.932 & \textbf{0.638} & \textbf{0.799} & 0.893 & 0.769 & 61.68 & 1.69\\
 & iTransformer & 0.144 & 0.201 & 0.876 & 0.636 & 0.798 & 0.891 & 0.760 & 48.48 & 0.88\\
\midrule

\multirow{9}{*}{168}
 & DLinear      & 0.139 & 0.199 & 1.059 & 0.647 & 0.804 & 0.881 & 0.747 & \textbf{20.65} & \textbf{0.37} \\
 & LightTS      & 0.135 & 0.194 & 0.910 & 0.659 & 0.812 & 0.879 & 0.757 & 36.93 & 0.84\\
 & TSMixer      & 0.134 & 0.195 & 0.966 & 0.661 & 0.813 & 0.882 & 0.760 & 33.46 & 0.98\\
 & FiLM         & 0.150 & 0.207 & 0.912 & 0.620 & 0.788 & 0.873 & 0.777 & 344.58 & 14.42\\
 & SCINet       & 0.138 & 0.197 & 0.855 & 0.650 & 0.806 & 0.882 & 0.855 & 576.19 & 14.27\\
 & TiDE         & 0.139 & 0.198 & 0.869 & 0.647 & 0.804 & 0.882 & \textbf{0.743} & 42.07 & 1.33\\
 & Crossformer  & 0.136 & 0.194 & 0.835 & 0.656 & 0.810 & 0.883 & 0.760 & 379.45 & 8.91\\
 & PatchTST     & \textbf{0.133} & 0.193 & 0.864 & \textbf{0.663} & \textbf{0.814} & \textbf{0.885} & 0.771 & 67.86 & 3.03\\
 & iTransformer & 0.135 & \textbf{0.192} & \textbf{0.830} & 0.659 & 0.812 & 0.881 & 0.772 & 48.61 & 1.21 \\
\midrule

\multirow{9}{*}{336}
 & DLinear      & 0.135 & 0.196 & 0.925 & 0.657 & 0.811 & 0.877 & 0.856 & 20.56 & \textbf{0.23}\\
 & LightTS      & 0.131 & 0.192 & 0.942 & 0.667 & 0.817 & 0.875 & 0.855 & \underline{\textbf{20.56}} & 1.30\\
 & TSMixer      & 0.130 & 0.190 & 0.959 & 0.670 & 0.818 & 0.874 & 0.854 & 34.56 & 1.03\\
 & FiLM         & 0.150 & 0.208 & 0.936 & 0.619 & 0.787 & 0.861 & 0.894 & 337.53 & 15.59\\
 & SCINet       & 0.134 & 0.195 & 0.879 & 0.660 & 0.812 & 0.876 & 0.851 & 895.46 & 16.42\\
 & TiDE         & 0.135 & 0.195 & 0.882 & 0.657 & 0.811 & 0.877 & 0.854 & 45.76 & 2.27\\
 & Crossformer  & 0.132 & 0.192 & 0.834 & 0.665 & 0.816 & \textbf{0.878} & 0.853 & 433.34 & 7.06\\
 & PatchTST     & \textbf{0.130} & \underline{\textbf{0.189}} & \underline{\textbf{0.816}} & \textbf{0.672} & \textbf{0.819} & 0.875 & 0.846 & 80.72 & 1.77\\
 & iTransformer & 0.131 & 0.190 & 0.829 & 0.667 & 0.817 & 0.877 & \textbf{0.851} & 47.45 & 1.21\\
\midrule

\multirow{9}{*}{672}
 & DLinear      & 0.134 & 0.195 & 0.946 & 0.660 & 0.813 & 0.890 & 0.693 & \textbf{21.31} & \underline{\textbf{0.20}}\\
 & LightTS      & 0.131 & 0.192 & 0.924 & 0.667 & 0.817 & 0.892 & 0.677 & 36.86 & 0.60\\
 & TSMixer      & 0.130 & 0.192 & 0.998 & 0.669 & 0.818 & 0.897 & \underline{\textbf{0.673}} & 41.20 & 1.01\\
 & FiLM         & 0.150 & 0.208 & 0.951 & 0.619 & 0.787 & 0.876 & 0.754 & 325.80 & 14.97\\
 & SCINet       & 0.133 & 0.194 & 0.898 & 0.663 & 0.814 & 0.889 & 0.694 & 1402.48 & 21.77\\
 & TiDE         & 0.134 & 0.194 & 0.892 & 0.661 & 0.813 & 0.890 & 0.695 & 42.89 & 2.28\\
 & Crossformer  & 0.131 & 0.190 & 0.849 & 0.669 & 0.818 & 0.892 & 0.691 & 447.89 & 9.40\\
 & PatchTST     & \underline{\textbf{0.128}} & \textbf{0.190} & 0.897 & \underline{\textbf{0.674}} & \underline{\textbf{0.821}} & \underline{\textbf{0.897}} & 0.689 & 81.99 & 1.97\\
 & iTransformer & 0.130 & 0.191 & \textbf{0.819} & 0.669 & 0.818 & 0.896 & 0.686 & 47.27 & 2.13\\
\bottomrule
\end{tabular}

\end{table}

\begin{table}
\centering
\caption{Forecasting performance on the LCL dataset across different historical context lengths ($s$, in hours) for \texttt{pred\_len} = 96. MAPE values are reported in units of $\times 10^{3}$. Here, $r$ denotes the Pearson correlation coefficient, $\cos$ the cosine similarity, $FD$ the Fréchet distance, and $tt$ and $it$ the training and inference time in minutes and milliseconds, respectively. The best result per context length is highlighted in \textbf{bold}, and the overall best result is emphasized in \underline{\textbf{bold and underlined}}.}
\label{tab:lcl-96}
\begin{tabular}{C{0.7cm} C{1.8cm} C{1.1cm} C{1.1cm} C{1.3cm} C{1.1cm} C{1.1cm} C{1.1cm} C{1.1cm} C{1.0cm} C{0.7cm}}
\toprule
$s$ & Model & MSE $\downarrow$ & MAE $\downarrow$ & MAPE $\downarrow$ & $R^2$ $\uparrow$ & $r$ $\uparrow$ & $\cos$ $\uparrow$ & FD $\downarrow$ & $tt$ $\downarrow$ & $it$ $\downarrow$ \\
\midrule

% ---------------- 24 ----------------
\multirow{9}{*}{24}
 & DLinear      & 0.192 & 0.241 & 1.603 & 0.520 & 0.722 & 0.827 & 1.254 & \textbf{21.88} & \underline{\textbf{0.24}} \\
 & LightTS      & 0.180 & 0.231 & 1.733 & 0.550 & 0.742 & 0.835 & 1.253 & 37.09 & 0.80 \\
 & TSMixer      & 0.181 & 0.230 & 1.530 & 0.546 & 0.740 & 0.835 & 1.265 & 31.23 & 0.97 \\
 & FiLM         & 0.193 & 0.233 & \textbf{0.961} & 0.516 & 0.720 & 0.826 & 1.262 & 196.28 & 7.47 \\
 & SCINet       & 0.189 & 0.229 & 0.968 & 0.526 & 0.727 & 0.833 & 1.251 & 310.05 & 12.19 \\
 & TiDE         & 0.193 & 0.233 & 0.963 & 0.517 & 0.720 & 0.826 & 1.260 & 43.24 & 1.33 \\
 & Crossformer  & \textbf{0.179} & 0.227 & 1.498 & \textbf{0.553} & \textbf{0.744} & 0.836 & 1.267 & 307.14 & 9.03 \\
 & PatchTST     & 0.181 & \textbf{0.226} & 1.079 & 0.547 & 0.741 & \textbf{0.836} & 1.299 & 50.39 & 1.69 \\
 & iTransformer & 0.179 & 0.227 & 1.563 & 0.552 & 0.743 & 0.835 & \textbf{1.251} & 48.62 & 1.20 \\
\midrule

% ---------------- 96 ----------------
\multirow{9}{*}{96}
 & DLinear      & 0.160 & 0.215 & 1.257 & 0.600 & 0.775 & 0.847 & 1.167 & \textbf{21.55} & \textbf{0.24} \\
 & LightTS      & 0.153 & 0.211 & 1.139 & 0.616 & 0.785 & 0.849 & 1.188 & 38.71 & 0.83 \\
 & TSMixer      & 0.153 & 0.211 & 1.231 & 0.616 & 0.785 & 0.852 & 1.185 & 33.10 & 0.95 \\
 & FiLM         & 0.160 & 0.212 & \textbf{0.945} & 0.599 & 0.774 & 0.846 & 1.172 & 510.94 & 23.00 \\
 & SCINet       & 0.158 & 0.212 & 0.983 & 0.605 & 0.778 & 0.849 & 1.157 & 382.49 & 13.18 \\
 & TiDE         & 0.160 & 0.212 & 0.947 & 0.599 & 0.774 & 0.846 & \textbf{1.167} & 44.53 & 1.33 \\
 & Crossformer  & 0.154 & 0.209 & 1.093 & 0.613 & 0.783 & 0.849 & 1.181 & 320.07 & 9.06 \\
 & PatchTST     & \textbf{0.153} & \textbf{0.208} & 1.076 & 0.615 & 0.785 & \textbf{0.855} & 1.190 & 52.52 & 1.70 \\
 & iTransformer & 0.153 & 0.209 & 1.084 & \textbf{0.617} & \textbf{0.786} & 0.851 & 1.197 & 47.88 & 1.20 \\
\midrule

% ---------------- 168 ----------------
\multirow{9}{*}{168}
 & DLinear      & 0.149 & 0.206 & 1.139 & 0.627 & 0.792 & 0.849 & 1.090 & \underline{\textbf{20.44}} & \textbf{0.24} \\
 & LightTS      & 0.145 & 0.203 & 1.058 & 0.637 & 0.798 & 0.851 & 1.118 & 38.46 & 1.31 \\
 & TSMixer      & 0.144 & 0.202 & 1.099 & 0.638 & 0.799 & 0.851 & 1.130 & 33.15 & 1.03 \\
 & FiLM         & 0.149 & 0.205 & 0.945 & 0.625 & 0.791 & 0.847 & 1.099 & 743.67 & 33.28 \\
 & SCINet       & 0.148 & 0.204 & 0.955 & 0.630 & 0.794 & 0.848 & 1.100 & 565.89 & 14.25 \\
 & TiDE         & 0.149 & 0.204 & \underline{\textbf{0.944}} & 0.626 & 0.792 & 0.847 & \textbf{1.090} & 41.76 & 1.33 \\
 & Crossformer  & 0.144 & 0.202 & 0.996 & \textbf{0.638} & \textbf{0.799} & \textbf{0.854} & 1.112 & 339.76 & 9.13 \\
 & PatchTST     & 0.145 & 0.202 & 0.992 & 0.637 & 0.798 & 0.852 & 1.123 & 65.45 & 1.79 \\
 & iTransformer & \textbf{0.146} & \textbf{0.201} & 0.996 & 0.634 & 0.797 & 0.852 & 1.139 & 47.84 & 1.22 \\
\midrule

% ---------------- 336 ----------------
\multirow{9}{*}{336}
 & DLinear      & 0.144 & 0.202 & 1.086 & 0.638 & 0.799 & 0.855 & 1.116 & \textbf{21.56} & \textbf{0.24} \\
 & LightTS      & 0.143 & 0.203 & 1.055 & 0.642 & 0.802 & 0.854 & 1.126 & 36.19 & 1.30 \\
 & TSMixer      & 0.141 & \underline{\textbf{0.198}} & 1.032 & 0.646 & 0.804 & 0.855 & 1.162 & 33.78 & 1.00 \\
 & FiLM         & 0.146 & 0.202 & 0.956 & 0.635 & 0.797 & 0.857 & 1.125 & 1007.12 & 47.24 \\
 & SCINet       & 0.144 & 0.201 & 0.975 & 0.640 & 0.800 & 0.856 & 1.123 & 853.27 & 16.37 \\
 & TiDE         & 0.144 & 0.201 & \textbf{0.953} & 0.638 & 0.799 & 0.856 & \textbf{1.116} & 36.70 & 2.19 \\
 & Crossformer  & \textbf{0.141} & 0.199 & 0.996 & \textbf{0.646} & \textbf{0.804} & \textbf{0.859} & 1.150 & 345.71 & 9.24 \\
 & PatchTST     & 0.142 & 0.202 & 1.023 & 0.644 & 0.802 & 0.856 & 1.169 & 69.21 & 2.32 \\
 & iTransformer & 0.142 & 0.200 & 0.994 & 0.643 & 0.802 & 0.854 & 1.156 & 46.79 & 1.20 \\
\midrule

% ---------------- 672 ----------------
\multirow{9}{*}{672}
 & DLinear      & 0.144 & 0.203 & 1.030 & 0.638 & 0.799 & 0.861 & 1.000 & \textbf{21.09} & \textbf{0.38} \\
 & LightTS      & 0.145 & 0.204 & 1.132 & 0.637 & 0.798 & 0.859 & \underline{\textbf{0.994}} & 30.65 & 0.60 \\
 & TSMixer      & 0.142 & 0.202 & 1.093 & 0.644 & 0.803 & 0.859 & 1.026 & 39.00 & 0.99 \\
 & FiLM         & 0.146 & 0.203 & 0.973 & 0.633 & 0.796 & 0.858 & 1.019 & 1028.61 & 37.47 \\
 & SCINet       & 0.143 & 0.201 & 0.985 & 0.642 & 0.801 & 0.860 & 1.000 & 1385.01 & 21.56 \\
 & TiDE         & 0.144 & 0.201 & \textbf{0.966} & 0.639 & 0.800 & 0.861 & 1.009 & 46.36 & 1.46 \\
 & Crossformer  & \underline{\textbf{0.141}} & \textbf{0.199} & 0.995 & \underline{\textbf{0.646}} & \underline{\textbf{0.804}} & 0.862 & 1.026 & 392.13 & 7.25 \\
 & PatchTST     & 0.142 & 0.204 & 1.071 & 0.643 & 0.802 & 0.862 & 1.027 & 70.25 & 2.43 \\
 & iTransformer & 0.144 & 0.201 & 0.998 & 0.638 & 0.799 & \underline{\textbf{0.863}} & 1.015 & 46.53 & 0.89 \\
\bottomrule
\end{tabular}
\end{table}

\begin{table}
\centering
\caption{Forecasting performance on the LCL dataset across different historical context lengths ($s$, in hours) for \texttt{pred\_len} = 168. MAPE values are reported in units of $\times 10^{3}$. Here, $r$ denotes the Pearson correlation coefficient, $\cos$ the cosine similarity, $FD$ the Fréchet distance, and $tt$ and $it$ the training and inference time in minutes and milliseconds, respectively. The best result per context length is highlighted in \textbf{bold}, and the overall best result is emphasized in \underline{\textbf{bold and underlined}}.}
\label{tab:lcl-168}
\begin{tabular}{C{0.7cm} C{1.8cm} C{1.1cm} C{1.1cm} C{1.3cm} C{1.1cm} C{1.1cm} C{1.1cm} C{1.1cm} C{1.0cm} C{0.7cm}}
\toprule
$s$ & Model & MSE $\downarrow$ & MAE $\downarrow$ & MAPE $\downarrow$ & $R^2$ $\uparrow$ & $r$ $\uparrow$ & $\cos$ $\uparrow$ & FD $\downarrow$ & $tt$ $\downarrow$ & $it$ $\downarrow$ \\
\midrule

% ---------------- 24 ----------------
\multirow{9}{*}{24}
 & DLinear      & 0.195 & 0.244 & 1.687 & 0.517 & 0.719 & 0.824 & 1.443 & \textbf{22.06} & \underline{\textbf{0.24}} \\
 & LightTS      & 0.183 & 0.232 & 1.650 & 0.546 & 0.739 & 0.832 & 1.453 & 36.81 & 0.79 \\
 & TSMixer      & 0.185 & 0.233 & 1.627 & 0.540 & 0.736 & 0.832 & 1.480 & 30.84 & 0.98 \\
 & FiLM         & 0.196 & 0.235 & \textbf{0.982} & 0.513 & 0.718 & 0.825 & 1.448 & 195.83 & 7.47 \\
 & SCINet       & 0.192 & 0.232 & 1.006 & 0.523 & 0.725 & 0.829 & 1.456 & 309.90 & 12.11 \\
 & TiDE         & 0.196 & 0.235 & 0.982 & 0.513 & 0.718 & 0.825 & 1.450 & 44.40 & 1.31 \\
 & Crossformer  & \textbf{0.182} & \textbf{0.229} & 1.592 & \textbf{0.549} & \textbf{0.741} & \textbf{0.834} & 1.436 & 247.66 & 8.99 \\
 & PatchTST     & 0.184 & 0.229 & 1.139 & 0.544 & 0.738 & 0.832 & 1.478 & 46.65 & 1.72 \\
 & iTransformer & 0.182 & 0.231 & 1.644 & 0.549 & 0.741 & 0.833 & \textbf{1.408} & 48.34 & 1.20 \\
\midrule

% ---------------- 96 ----------------
\multirow{9}{*}{96}
 & DLinear      & 0.162 & 0.216 & 1.294 & 0.598 & 0.773 & 0.827 & \textbf{1.364} & \textbf{21.65} & \textbf{0.37} \\
 & LightTS      & 0.156 & 0.213 & 1.211 & 0.612 & 0.783 & 0.829 & 1.371 & 38.59 & 0.83 \\
 & TSMixer      & 0.156 & 0.215 & 1.372 & 0.612 & 0.782 & 0.829 & 1.394 & 32.76 & 1.09 \\
 & FiLM         & 0.163 & 0.213 & 0.957 & 0.596 & 0.773 & 0.826 & 1.367 & 490.90 & 15.71 \\
 & SCINet       & 0.160 & 0.212 & 0.978 & 0.602 & 0.776 & 0.827 & 1.384 & 370.66 & 16.60 \\
 & TiDE         & 0.162 & 0.213 & \underline{\textbf{0.955}} & 0.597 & 0.773 & 0.826 & 1.368 & 43.96 & 1.79 \\
 & Crossformer  & 0.157 & \textbf{0.211} & 1.044 & 0.610 & 0.781 & 0.830 & 1.376 & 251.37 & 9.92 \\
 & PatchTST     & 0.157 & 0.212 & 1.141 & 0.611 & 0.782 & 0.830 & 1.411 & 51.74 & 1.68 \\
 & iTransformer & \textbf{0.156} & 0.211 & 1.172 & \textbf{0.612} & \textbf{0.782} & \textbf{0.831} & 1.393 & 47.59 & 2.10 \\
\midrule

% ---------------- 168 ----------------
\multirow{9}{*}{168}
 & DLinear      & 0.153 & 0.209 & 1.233 & 0.619 & 0.787 & 0.835 & \textbf{1.307} & \textbf{20.63} & \textbf{0.24} \\
 & LightTS      & \textbf{0.150} & 0.207 & 1.155 & \textbf{0.627} & \textbf{0.792} & 0.838 & 1.337 & 38.31 & 0.98 \\
 & TSMixer      & 0.150 & 0.208 & 1.237 & 0.627 & 0.792 & 0.837 & 1.352 & 33.31 & 1.04 \\
 & FiLM         & 0.154 & 0.207 & \textbf{0.960} & 0.616 & 0.785 & 0.834 & 1.326 & 840.00 & 38.26 \\
 & SCINet       & 0.152 & 0.209 & 1.014 & 0.621 & 0.788 & 0.835 & 1.319 & 557.84 & 14.42 \\
 & TiDE         & 0.154 & 0.207 & 0.962 & 0.617 & 0.786 & 0.834 & 1.310 & 43.63 & 2.18 \\
 & Crossformer  & 0.151 & \textbf{0.206} & 1.124 & 0.625 & 0.791 & \textbf{0.838} & 1.328 & 247.69 & 9.13 \\
 & PatchTST     & 0.151 & 0.209 & 1.147 & 0.626 & 0.791 & 0.837 & 1.378 & 44.95 & 1.98 \\
 & iTransformer & 0.151 & 0.207 & 1.071 & 0.625 & 0.791 & 0.838 & 1.329 & 47.56 & 1.20 \\
\midrule

% ---------------- 336 ----------------
\multirow{9}{*}{336}
 & DLinear      & 0.149 & 0.206 & 1.175 & 0.629 & 0.793 & 0.846 & \textbf{1.306} & \underline{\textbf{20.46}} & \textbf{0.27} \\
 & LightTS      & 0.148 & 0.205 & 1.183 & 0.631 & 0.794 & 0.845 & 1.328 & 31.07 & 1.31 \\
 & TSMixer      & 0.146 & 0.202 & 1.080 & 0.635 & 0.797 & 0.846 & 1.341 & 33.56 & 0.93 \\
 & FiLM         & 0.151 & 0.205 & 0.976 & 0.625 & 0.791 & 0.846 & 1.336 & 1312.73 & 62.59 \\
 & SCINet       & 0.149 & 0.204 & 1.024 & 0.629 & 0.793 & 0.846 & 1.323 & 846.47 & 12.12 \\
 & TiDE         & 0.150 & 0.204 & \textbf{0.974} & 0.628 & 0.793 & 0.847 & 1.309 & 43.05 & 0.94 \\
 & Crossformer  & \underline{\textbf{0.146}} & 0.203 & 1.073 & \underline{\textbf{0.636}} & \underline{\textbf{0.798}} & 0.846 & 1.321 & 246.65 & 9.23 \\
 & PatchTST     & 0.148 & 0.205 & 1.064 & 0.632 & 0.795 & 0.846 & 1.418 & 61.85 & 2.93 \\
 & iTransformer & 0.148 & \underline{\textbf{0.202}} & 1.236 & 0.632 & 0.795 & \textbf{0.847} & 1.373 & 46.65 & 1.20 \\
\midrule

% ---------------- 672 ----------------
\multirow{9}{*}{672}
 & DLinear      & 0.149 & 0.205 & 1.110 & 0.629 & 0.793 & 0.852 & 1.287 & \textbf{21.27} & \textbf{0.26} \\
 & LightTS      & 0.150 & 0.208 & 1.259 & 0.625 & 0.791 & 0.851 & \underline{\textbf{1.282}} & 30.44 & 0.89 \\
 & TSMixer      & 0.148 & 0.206 & 1.169 & 0.631 & 0.795 & 0.852 & 1.299 & 34.08 & 1.01 \\
 & FiLM         & 0.151 & 0.206 & 1.002 & 0.624 & 0.790 & 0.849 & 1.315 & 1703.94 & 62.84 \\
 & SCINet       & 0.148 & 0.204 & 1.011 & 0.631 & 0.795 & 0.852 & 1.294 & 1376.08 & 21.19 \\
 & TiDE         & 0.149 & 0.204 & \textbf{0.995} & 0.629 & 0.793 & 0.852 & 1.288 & 45.71 & 1.32 \\
 & Crossformer  & 0.146 & \textbf{0.204} & 1.218 & \textbf{0.635} & \textbf{0.797} & \underline{\textbf{0.854}} & 1.315 & 256.17 & 9.33 \\
 & PatchTST     & \textbf{0.148} & 0.206 & 1.106 & 0.630 & 0.794 & 0.853 & 1.341 & 80.91 & 1.77 \\
 & iTransformer & 0.151 & 0.206 & 1.025 & 0.623 & 0.790 & 0.852 & 1.320 & 46.25 & 1.21 \\
\bottomrule
\end{tabular}
\end{table}

\subsection{Impact of Historical Context}
\label{sec:discussion-historical}

We assess the effect of historical context length (\texttt{seq\_len}) by sweeping the input window from 24 to 672 hours, with results reported in Tables~\ref{tab:lcl-24}--\ref{tab:lcl-168} for the LCL dataset. Across all models, longer historical context consistently improves performance in terms of MSE, MAE, MAPE, $R^2$, Pearson's $r$, and cosine similarity, as extended input windows enable models to capture periodic patterns and medium-term trends. However, the largest gains are observed when increasing \texttt{seq\_len} from 24 to 168 hours, after which improvements become marginal, suggesting that once daily and weekly seasonality are captured, additional context contributes limited new information and can even cause a plateau or slight degradation in some models, particularly LightTS and iTransformer at longer prediction horizons.

We also examine model-specific trends to understand how different architectures exploit additional context. Among linear and MLP-based models (DLinear, LightTS, and TSMixer), LightTS performs best at shorter context lengths while TSMixer takes over at longer ones (\texttt{seq\_len} $\geq$ 168), reflecting the benefit of its time- and feature-mixing mechanisms for capturing richer periodic structure. Among Transformer-based models, Crossformer leads at the shortest context length across all horizons, while PatchTST dominates at longer context lengths for \texttt{pred\_len} = 24 and Crossformer becomes the leading Transformer at longer contexts for \texttt{pred\_len} = 96 and 168, which suggests that cross-dimension attention mechanisms are increasingly beneficial as both context and horizon grow. With respect to structural similarity, TSMixer achieves the globally best Fréchet distance at \texttt{seq\_len} = 672 and \texttt{pred\_len} = 24, indicating robust preservation of global trajectory geometry due to its mixing mechanism capturing smooth periodic structure without introducing the local deviations that accumulate in attention-based architectures.

\subsection{Impact of Prediction Horizon}

We evaluate sensitivity to the prediction horizon (\texttt{pred\_len}) by considering horizons of 24, 96, and 168 hours (one day, four days, and one week), with results reported in Tables~\ref{tab:lcl-24}--\ref{tab:lcl-168} for the LCL dataset, where each table corresponds to a fixed horizon with \texttt{seq\_len} varied from 24 to 672 hours. As expected, performance degrades consistently as the prediction horizon increases across all \texttt{seq\_len} configurations, with MSE and MAE increasing and $R^2$, Pearson's $r$, and cosine similarity declining, since the widening temporal gap between observed inputs and forecast targets makes short-term dependencies less informative and forces predictions to rely on coarse-grained seasonal structure. Architectural rankings also shift with horizon: at \texttt{pred\_len} = 24, Crossformer leads at the shortest context (\texttt{seq\_len} = 24), while PatchTST dominates at medium and longer input windows, consistent with its patch-based attention effectively exploiting richer historical context; at \texttt{pred\_len} = 96, Crossformer again leads at the shortest and longest context lengths, while TSMixer and LightTS appear strong at medium input windows; and at \texttt{pred\_len} = 168, Crossformer leads at the shortest and longest context lengths, with LightTS performing best at intermediate windows, which indicates that the relative advantage of complex attention mechanisms diminishes at longer horizons where lighter architectures offer a competitive alternative. At the longest horizon, performance differences across architectures narrow considerably, suggesting that horizon-induced uncertainty limits the advantage conferred by architectural complexity. With respect to structural similarity, DLinear achieves the lowest Fréchet distance most consistently across horizons, particularly at \texttt{pred\_len} = 96 and 168, preserving trajectory-level shape more reliably than other models due to its trend-seasonal decomposition, which constrains predictions to smooth low-frequency dynamics. These results highlight that pointwise accuracy and structural fidelity are not equivalent objectives, and that model selection should reflect the criterion relevant to the deployment setting.

\begin{center}
\captionof{table}{Subgroup performance (MSE) across ACORN categories for \texttt{seq\_len} = 96 and \texttt{pred\_len} = 168. The lowest error per subgroup is highlighted in \textbf{bold}.}
\label{tab:acorn}
\begin{tabular}{|c|C{1.1cm}|C{1.1cm}|C{1.1cm}|C{1.1cm}|C{1.1cm}|C{1.1cm}|C{1.1cm}|C{1.3cm}|C{1.1cm}|}
\hline
& DLinear & LightTS & TSMixer & FiLM & SCINet & TiDE & CFormer & PatchTST & iTrans\\
\hline
\hline
Challenged & 0.192 & 0.174 & 0.175 & 0.192 & 0.191 & 0.191 & 0.164 & \textbf{0.134} & 0.155\\
\hline
\hline
Adversity & 0.108 & 0.105 & 0.105 & 0.108 & 0.107 & 0.108 & 0.105 & \textbf{0.104} & 0.105\\
\hline
\hline
Comfortable & 0.124 & 0.121 & 0.121 & 0.124 & 0.123 & 0.124 & 0.121 & 0.121 & \textbf{0.120}\\
\hline
\hline
Affluent & 0.231 & \textbf{0.221} & 0.222 & 0.233 & 0.228 & 0.233 & 0.224 & 0.224 & 0.223\\
\hline
\end{tabular}
\end{center}

\begin{center}
\captionof{table}{Subgroup performance (MSE) across building types for the Pecan Street dataset, \texttt{seq\_len} = 672 and \texttt{pred\_len} = 96. The lowest error per subgroup is highlighted in \textbf{bold}.}
\label{tab:pecan_subgroup}
\begin{tabular}{|c|C{1.1cm}|C{1.1cm}|C{1.1cm}|C{1.1cm}|C{1.1cm}|C{1.1cm}|C{1.1cm}|C{1.3cm}|C{1.1cm}|}
\hline
& DLinear & LightTS & TSMixer & FiLM & SCINet & TiDE & CFormer & PatchTST & iTrans\\
\hline
\hline
Apartment & 0.333 & 0.334 & 0.334 & 0.339 & 0.333 & 0.334 & \textbf{0.330} & 0.334 & 0.331\\
\hline
\hline
Single-Family & 0.530 & 0.528 & 0.527 & 0.539 & 0.528 & 0.529 & \textbf{0.519} & 0.529 & 0.527\\
\hline
\hline
Town Home & 1.302 & 1.304 & \textbf{1.280} & 1.307 & 1.298 & 1.305 & 1.403 & 1.327 & 1.344\\
\hline
\end{tabular}
\end{center}

\subsection{Subgroup Analysis}

\edit{In addition to aggregate accuracy, understanding whether forecasting performance is consistent across population segments matters. This is because uneven accuracy across socioeconomic or housing categories could translate into unequal outcomes in downstream applications such as demand-response eligibility or tariff design, making subgroup-level evaluation relevant to the fair deployment of these models, not just their overall performance.}

We further analyze the performance across geodemographic segments to assess whether forecasting accuracy varies across different groups of the population. In Table~\ref{tab:acorn}, we present MSE for each ACORN segment under a representative scenario of four days of historical context (\texttt{seq\_len} = 96) and one-week-ahead forecasts (\texttt{pred\_len} = 168). Although we focus on this scenario to present our results, we note that similar trends were observed across other combinations of context and horizon lengths. For the \emph{Adversity}, \emph{Comfortable}, and \emph{Affluent} segments, models perform very similarly to each other, with differences in MSE being marginal, which suggests that consumption behavior within these groups is sufficiently regular for all architectures to capture. In contrast, the \emph{Challenged} group shows a different pattern: PatchTST achieves the lowest MSE, with a larger gap over other models than observed in any other segment. This is interesting given that the \emph{Challenged} group is the most under-represented in the dataset (1.3\% of households), and that the second and third best performing models on this segment are Crossformer and iTransformer, respectively, \edit{which might tentatively suggest} that Transformer-based architectures are better suited to modeling irregular consumption patterns, especially in situations with less data. \edit{To quantify this, we computed household-level MSE per model for the Challenged group. With only 7 households, the standard error of the household-level mean is large for every model, ranging from 0.069 to 0.110. We note that even the full spread between the best- and worst-performing models corresponds to less than one standard error of the mean. This indicates that the observed ranking on this subgroup cannot be considered stable given the available sample size.}

We conduct the same analysis on the Pecan Street dataset, grouping households by building type instead (Table~\ref{tab:pecan_subgroup}). In this case, the pattern differs from LCL: although Crossformer achieves the lowest MSE on \emph{Apartment}, the smallest group (10\% of households), and TSMixer leads on \emph{Town Home} (20\% of households), the gap over other models is marginal in both cases, unlike the notable advantage observed for PatchTST on the \emph{Challenged} group in LCL. \edit{This suggests that while attention-based architectures can perform well on under-represented groups, their advantage is not a consistent, dataset-independent effect, and, where such an advantage does appear, its magnitude may relate to the degree of behavioral irregularity within a segment rather than to representation size alone.} We also note that, unlike the ACORN groups in LCL, where median consumption levels are comparable across categories (Figure~\ref{fig:lcl-b}), the building types in Pecan Street vary significantly in consumption magnitude, as discussed in Section~\ref{sec:datasets} (Figure~\ref{fig:pecan-b}). This explains why the absolute MSE values are substantially different across building types in Pecan Street.

\begin{center}
\captionof{table}{Vargha--Delaney $\hat{A}_{12}$ effect sizes for the LCL dataset at $\texttt{seq\_len}=168$ and $\texttt{pred\_len}=24$. Each entry $\hat{A}_{12}[i,j] = P(\text{row model error} > \text{column model error})$, where values above $0.5$ indicate that the row model tends to produce higher per-window errors.}
\label{tab:effect_sizes-lcl}
\setlength{\tabcolsep}{4pt}
\small
\begin{tabular}{lrrrrrrrrr}
\toprule
 & DLinear & LightTS & TSMixer & FiLM & SCINet & TiDE & CFormer & PatchTST & iTrans. \\
\midrule
DLinear & --- & 0.510 & 0.508 & 0.491 & 0.506 & 0.506 & 0.512 & 0.515 & 0.517 \\
LightTS & 0.490 & --- & 0.499 & 0.482 & 0.497 & 0.497 & 0.503 & 0.506 & 0.507 \\
TSMixer & 0.492 & 0.501 & --- & 0.483 & 0.498 & 0.498 & 0.504 & 0.507 & 0.508 \\
FiLM & 0.509 & 0.518 & 0.517 & --- & 0.515 & 0.514 & 0.521 & 0.523 & 0.525 \\
SCINet & 0.494 & 0.503 & 0.502 & 0.485 & --- & 0.500 & 0.506 & 0.509 & 0.510 \\
TiDE & 0.494 & 0.503 & 0.502 & 0.486 & 0.500 & --- & 0.506 & 0.509 & 0.510 \\
CFormer & 0.488 & 0.497 & 0.496 & 0.479 & 0.494 & 0.494 & --- & 0.503 & 0.504 \\
PatchTST & 0.485 & 0.494 & 0.493 & 0.477 & 0.491 & 0.491 & 0.497 & --- & 0.501 \\
iTrans. & 0.483 & 0.493 & 0.492 & 0.475 & 0.490 & 0.490 & 0.496 & 0.499 & --- \\
\bottomrule
\end{tabular}
\end{center}

\begin{center}
\captionof{table}{Vargha--Delaney $\hat{A}_{12}$ effect sizes for the Pecan Street dataset at $\texttt{seq\_len}=96$ and $\texttt{pred\_len}=672$. Each entry $\hat{A}_{12}[i,j] = P(\text{row model error} > \text{column model error})$, where values above $0.5$ indicate that the row model tends to produce higher per-window errors.}
\label{tab:effect_sizes-pecan}
\setlength{\tabcolsep}{3pt}
\footnotesize
\begin{tabular}{lrrrrrrrrr}
\toprule
 & DLinear & LightTS & TSMixer & FiLM & SCINet & TiDE & CFormer & PatchTST & iTrans. \\
\midrule
DLinear & --- & 0.546 & 0.559 & 0.628 & 0.621 & 0.628 & 0.516 & 0.626 & 0.542 \\
LightTS & 0.454 & --- & 0.514 & 0.590 & 0.583 & 0.590 & 0.468 & 0.587 & 0.496 \\
TSMixer & 0.441 & 0.486 & --- & 0.578 & 0.570 & 0.578 & 0.455 & 0.574 & 0.482 \\
FiLM & 0.372 & 0.410 & 0.422 & --- & 0.493 & 0.500 & 0.383 & 0.495 & 0.406 \\
SCINet & 0.379 & 0.417 & 0.430 & 0.507 & --- & 0.508 & 0.389 & 0.502 & 0.414 \\
TiDE & 0.372 & 0.410 & 0.422 & 0.500 & 0.492 & --- & 0.382 & 0.495 & 0.406 \\
CFormer & 0.484 & 0.532 & 0.545 & 0.617 & 0.611 & 0.618 & --- & 0.615 & 0.528 \\
PatchTST & 0.374 & 0.413 & 0.426 & 0.505 & 0.498 & 0.505 & 0.385 & --- & 0.409 \\
iTrans. & 0.458 & 0.504 & 0.518 & 0.594 & 0.586 & 0.594 & 0.472 & 0.591 & --- \\
\bottomrule
\end{tabular}
\end{center}

\subsection{Statistical Analysis}

We have, so far, investigated model rankings across prediction horizons and sequence lengths. To assess whether these differences among models are statistically significant, we conduct a series of statistical tests. We use MAE as the error metric for all statistical tests since it provides a scale-interpretable and outlier-robust measure of forecast accuracy. We first apply the Friedman test, which is a non-parametric test that ranks models within each forecast window and tests whether the resulting average ranks are consistent with the null hypothesis that all models perform equally. A significant result indicates that at least one model is ranked differently from the others. For both datasets, the test rejected the null hypothesis across all cases ($p < 0.001$ in every case), confirming that the observed model rankings do not occur by chance. We also conduct a post-hoc analysis using Holm--Bonferroni-corrected pairwise Wilcoxon signed-rank tests on 5\,000 subsampled windows per configuration. On both datasets, DLinear is significantly outperformed by all other models in every configuration. For the remaining models, the significance pattern is mostly configuration-dependent on LCL, while on Pecan Street almost all pairs are significant at longer horizons.

Given the large number of overlapping test windows, statistical significance is expected even for negligible differences. We therefore compute the Vargha--Delaney $\hat{A}_{12}$ effect size for each model pair to quantify the practical magnitude of the detected differences, as reported in Tables~\ref{tab:effect_sizes-lcl} and~\ref{tab:effect_sizes-pecan}. $\hat{A}_{12}[i,j]$ estimates the probability that model $i$ produces a higher MAE than model $j$ on a randomly selected forecast window; a value of $0.5$ indicates no difference, while the assumed thresholds for small, medium, and large effects are $0.56$, $0.64$, and $0.71$, respectively. The two datasets yield different findings. On the LCL dataset, $\hat{A}_{12}$ values fall within $[0.467, 0.533]$ across all 15 configurations and all 36 model pairs, uniformly below the threshold for even a small effect, which shows that the nine deep learning models perform near-equivalently regardless of input length or forecast horizon. On the Pecan Street dataset, however, the pattern is more complex and reveals a clear horizon-dependent structure: at the shortest horizon ($\texttt{pred\_len} = 96$), effects are small, but they grow with forecast horizon. Several pairs, most consistently DLinear and Crossformer against models such as TiDE, PatchTST, SCINet, and FiLM, cross the small-effect threshold at longer horizons, demonstrating that architectural differences begin to matter practically as the forecasting task becomes harder. Importantly, this practical equivalence holds uniformly across all input lengths for a given horizon. This means that it is the forecast horizon, not the input sequence length, that determines whether model choice is meaningful on the Pecan Street dataset. These results establish that model selection is unlikely to be the primary driver of forecast accuracy at short horizons on residential energy data, but that architectural differences become relevant at longer horizons, especially on more diverse household populations.

\edit{To assess whether the pairwise differences are robust to within-household dependence among overlapping test windows, we conducted a household-level cluster bootstrap across all 15 LCL configurations, re-sampling households rather than windows (2\,000 iterations per configuration). The results show that of 540 pairwise model comparisons (36 pairs $\times$ 15 configurations), 453 (83.9\%) remain statistically significant under this clustered analysis. Seven pairs, including DLinear vs. TSMixer, TiDE, SCINet, and LightTS, are significant in every configuration tested. The least robust pairs are TSMixer vs. SCINet (non-significant in 7 of 15 configurations) and LightTS vs. SCINet (non-significant in 6 of 15 configurations). A robustness check against abnormal consumption spikes, included in the Supplementary Material, confirms that all nine models degrade similarly under such conditions.}

\section{Conclusion and Practical Implications}

In this paper, we benchmarked nine different deep learning forecasting models for forecasting residential energy consumption, evaluated on two high-resolution energy datasets. Our results carry several important implications for real-world forecasting systems. First, we showed that deep learning models consistently outperform classical baselines such as na\"ive seasonal and SARIMA in overall forecasting accuracy. Second, we demonstrated that simpler deep learning models are fairly comparable to more complex and computationally expensive Transformer-based models, meaning that simpler models may be a better choice for forecasting across multiple households due to their lower computational expense. \edit{For example, at \texttt{seq\_len} = 672 and \texttt{pred\_len} = 24 on the LCL dataset, TSMixer achieves an MSE of 0.130, within 1.5\% of PatchTST's best MSE of 0.128, while requiring roughly half the training time.} This is further supported by our statistical analysis, which shows that on the LCL dataset, differences between model architectures are statistically significant but practically negligible, whereas on the Pecan Street dataset, architectural differences become meaningful at longer forecasting horizons. Third, we showed that forecasting performance is only improved by increasing the historical window of data input into the forecasting model up to the point that dominant seasonal patterns, such as a weekly cycle, are captured, after which additional historical data does little to improve forecasting performance. This implies that for forecasting across individual households, moderate historical input windows may be sufficient. Fourth, we showcased that forecasting accuracy is negatively related to the length of the forecast horizon, which highlights the importance of selecting forecasting models and techniques that are appropriate for the forecasting horizon of a given application. 
\edit{Finally, we found that population differences had limited impact on model choice overall. The apparent Transformer-based advantage on under-represented LCL groups did not hold up statistically and did not generalize across datasets.} \edit{These findings also carry practical implications for deployment. Since lightweight models such as TSMixer and LightTS achieve accuracy competitive with more complex architectures at substantially lower training cost, utilities forecasting across large meter populations do not necessarily need to utilize the most computationally expensive models. We also note that seasonal drift in consumption patterns may require periodic retraining, which is an operational cost that is not captured by a single train/test split and worth accounting for in deployment planning.}

%Finally, we revealed that while forecasting performance is generally consistent across population segments, the advantage of Transformer-based models for under-represented groups observed on the LCL dataset does not consistently generalize across datasets, indicating that the impact of population differences on model selection is itself dataset-dependent and should be assessed empirically rather than assumed.

\section*{Acknowledgments}
This research was funded by the Swedish Energy Agency (Energimyndigheten) under grant
P2024-01187, within the TERMO program (2024-2028).
\printcredits

\bibliographystyle{cas-model2-names}

\bibliography{references}



\end{document}